\documentclass[11pt]{article}

\usepackage[final]{acl}
\usepackage{booktabs}
\usepackage{tabularx}
\usepackage{times}
\usepackage{latexsym}
\usepackage{array}
\usepackage[T1]{fontenc}

\usepackage[utf8]{inputenc}

\usepackage{microtype}

\usepackage{inconsolata}

\usepackage{graphicx}

\usepackage{booktabs}
\usepackage{amssymb,amsmath}
\usepackage{lscape}

\usepackage{xcolor}

\usepackage{forest}
\usepackage[dvipsnames]{xcolor}

\usepackage{pifont}
\usepackage{tikz}

\usepackage{fdsymbol}

\usepackage{longtable}

\usepackage{pdfpages}
      
\newcommand{\xmark}{\ding{55}}

\title{Challenges and Recommendations for LLM-as-a-Judge in Multilingual Settings and for Low-Resource Languages}

\author{
    A.Seza Doğruöz\textsuperscript{$\varheartsuit$} \quad  
    Xixian Liao\textsuperscript{$\clubsuit$} \quad 
    Verena Blaschke\textsuperscript{$\spadesuit$} \quad   \\
    \textbf{Jakob Prange\textsuperscript{$\vardiamondsuit,\diamondsuit$}} \quad
    \textbf{Senyu Li\textsuperscript{$\heartsuit$,$\largeblacksquare$}} \quad   
    \textbf{David Ifeoluwa Adelani\textsuperscript{$\heartsuit$,$\largeblacksquare$,$\maltese$}} 
    \\ \\
    \textsuperscript{$\varheartsuit$}LT3, IDLab, Universiteit Gent, 
    \textsuperscript{$\clubsuit$}Barcelona Supercomputing Center,\\
    \textsuperscript{$\spadesuit$}LMU Munich \& Munich Center for Machine Learning,
    \\
     \textsuperscript{$\vardiamondsuit$}German Center for Addiction Research in Childhood and Adolescence,
    \\
    \textsuperscript{$\diamondsuit$}University Medical Center Hamburg-Eppendorf,
    \\
    \textsuperscript{$\heartsuit$}Mila - Quebec AI Institute, 
    \textsuperscript{$\largeblacksquare$}McGill University, 
    \textsuperscript{$\maltese$}Canada CIFAR AI Chair\\
    \texttt{as.dogruoz@ugent.be}
}

\begin{document}
\maketitle

\begin{abstract}
LLM-as-a-Judge has become the dominant evaluation paradigm for many natural language generation tasks (albeit mostly in English) due to shortcomings of conventional metrics and high correlations with human judgment. 
There are now attempts to extend LLM-as-a-Judge to multilingual settings including low-resource languages. However, LLMs have limited proficiency in low-resource languages, and there is often no adequate human validation in these settings. %
To highlight the scope of the problem and current practices, we explore the use of LLM-as-a-Judge evaluators in ACL Anthology papers focusing on multilingual settings and low-resource languages across a diverse set of tasks. Out of 650 papers mentioning LLM-as-a-judge, only 33 of them focus on low-resource or multilingual settings. Our in-depth analysis of these papers indicates inconsistent evaluation outcomes, a tendency to overtrust LLM judgments in multilingual settings, and the widespread reliance on a single judge model per study. To help the NLP community further, we conclude with a checklist of recommendations about how to use LLM-as-a-Judge in multilingual and low-resource settings.

\end{abstract}

\section{Introduction}

Until recently, evaluation in Natural Language Processing (NLP) systems has mainly been performed by humans, with advantages (e.g., reliability, faithfulness, interpretability, quality control) and disadvantages like time-consuming procedures and financial costs \citep[][\textit{inter alia}]{muhammad-etal-2025-brighter,hellwig-etal-2025-still,wiechetek-etal-2025-create,rouzegar-makrehchi-2024-enhancing}. 

Large language models (LLMs) have become central to the development of NLP systems across a diverse range of tasks~\cite{achiam2023gpt,liu2024deepseek,team2024gemini}. In addition to directly performing  natural language generation (NLG) and understanding tasks, LLMs are increasingly used for evaluating the outputs of other language models~\cite{zheng-etal-2023-judging,adlakha-etal-2024-evaluating} as well. This paradigm is known as \textbf{LLM-as-a-Judge} and it profits from many features of state-of-the-art generative language modeling. These features include instruction following, multi-step reasoning, ease of use through chat conversation capabilities, high linguistic proficiency in high-resource languages, and the ability to 
generate explanations alongside outputs and judgments.

While human experts are still the upper bound for overall evaluation quality and trust, LLM-as-a-Judge is considered to be easier, cheaper and faster to scale. However, there is limited research about whether LLM judges deliver reliable and trustworthy evaluations that correlate strongly with human judges. These questions have previously been raised by \citet{zheng-etal-2023-judging}, who also coined the term {LLM-as-a-Judge}, and subsequently by \citet{chen-etal-2024-humans} and \citet{bavaresco-etal-2025-llms}, \textit{inter alia}. 
While some of these studies observe high correlations between LLM and human judges, their focus is limited to English. 

Nevertheless, the 
current trend of using LLM-as-a-Judge has been extended and scaled to languages other than English, including low-resource languages (LRLs). Although some studies have highlighted low reliability in such settings~\cite{hada-etal-2024-metal, hada-etal-2024-large}, LLM-as-a-Judge is becoming a mainstream evaluation method due to (1)~its perceived high correlation with human judgments (which is often only validated for a small number of languages); (2)~fast and cheap evaluation; 
and (3)~limitations and biases of traditional metrics, especially ones that are reference-based. 
As a result, LLM-as-a-Judge has become a predominant evaluation paradigm for NLP tasks (e.g., question answering and instruction following), and is now adopted across additional modalities, including visual question answering \citep{chen2024mllm} and AudioQA~\cite{ivry2026lalm}.

However, several critical issues arise when LLM-as-a-Judge systems are applied without sufficient caution across languages, including: (1)~inconsistent evaluation outcomes depending on the prompt language, with performance often overestimated for low-resource languages, (2)~inaccurate performance estimation due to the limited proficiency or inadequate understanding of outputs expressed in low-resource languages, and (3)~the widespread reliance on a single judge for evaluation, without considering multi-judge or ensemble-based assessment. At the same time, (4)~the tendency to overtrust LLM judgments across settings;\footnote{Overtrusting LLM judgments becomes particularly insidious when what is judged are human-generated outputs and annotations. For example, \citet{belay2026trust} use LLM-as-a-Judge to audit the quality of human translation for LRLs even though related work shows this is unreliable without in-context learning examples~\cite{li-etal-2025-ssa}. Similarly, \citet{cheng_etal_2026_sycophantic} find that, in difficult social scenarios, humans trust LLMs that evaluate their own stances favourably, even when they are clearly in the wrong.} clashes with (5)~the extremely high requirements for reliability when evaluating high-risk outputs involving safety, fairness, or cultural biases.

Previous work on multilingual LLM-as-a-Judge shows that judgment consistency is particularly weak for low-resource languages, raising concerns that LLM judges may be less reliable in the lower-resource settings where automatic evaluation is especially tempting due to the scarcity of expert annotators~\citep{fu-liu-2025-reliable, akinode2026tukabenchculturallygroundedjailbreak}.  Yet, we lack a systematic analysis about to what extent current multilingual and low-resource uses of LLM-as-a-Judge are reliable (i.e., whether judges are validated in the target languages, whether low-resource languages receive direct human or gold-label checks, and whether studies avoid over-reliance on single general-purpose judges). Our paper addresses this gap in the literature through a systematic survey of 33 in-scope papers retrieved from multilingual and low-resource LLM-as-a-Judge search criteria, analyzing how judges are deployed and validated across languages, tasks, and model families, and offering recommendations for more reliable cross-lingual evaluation.
Table~\ref{tab:taxonomy-summary} provides a compact preview of the five 
dimensions we used to characterize the surveyed papers. A detailed paper-level 
breakdown is provided in Figure~\ref{fig:taxonomy}.
\begin{table}[t]
\centering
\small
\begin{tabularx}{\columnwidth}{@{}p{0.27\columnwidth}X@{}}
\toprule
\textbf{Dimension} & \textbf{Categories} \\
\midrule
Judging paradigm 
& Direct scoring; pairwise/ranking; multi-step/reasoning; 
multi-dimensional; multi-agent/ensemble \\

Downstream task 
& MT; summarization; safety/toxicity; instruction following; 
QA/reasoning; RAG; other \\

Judge model 
& Closed-source; open-source; specialized/trained-as-judge \\

Language coverage 
& Monolingual; moderately multilingual (2--10); 
massively multilingual (10+) \\

Judge evaluation 
& Human/gold agreement; internal consistency/robustness; 
no direct evaluation \\
\bottomrule
\end{tabularx}
\caption{Summary of the five dimensions used in our taxonomy. 
Figure~\ref{fig:taxonomy} provides the full paper-level breakdown.}
\label{tab:taxonomy-summary}
\end{table}

\section{Related Work}

The LLM-as-a-Judge paradigm was introduced by \citet{zheng-etal-2023-judging}, who proposed MT-Bench and Chatbot Arena to evaluate chatbot alignment with human judgments. Their findings (based on evaluation in English) show that frontier LLMs (e.g., GPT-4) match human judgment at over 80\% agreement. They also identified key failure modes such as \textit{position bias} (sensitivity to the order in which candidate responses are presented), \textit{verbosity bias} (preference for longer responses regardless of quality), and \textit{self-enhancement bias} (tendency of a model to favor its own outputs). 

Building on this foundation, several \textbf{surveys} have since synthesized the growing body of work on LLM-based evaluation across NLP tasks. 
For example, \citet{gu-2024-survey} provide a comprehensive overview about how to build reliable LLM-as-a-Judge systems, covering bias mitigation, consistency improvement, and prompt design strategies and
\citet{li-etal-2024-llms-judges} analyze the LLMs-as-judges paradigm from five perspectives (e.g., functionality, methodology, application, meta-evaluation, and limitations)
Similarly, \citet{li-etal-2025-generation} organize the literature around LLM-as-a-Judge in three dimensions: \textit{what} to judge (quality attributes such as helpfulness, safety, and reliability), \textit{how} to judge (tuning and prompting strategies), and \textit{how to benchmark} LLM judges (categorizing benchmarks for LLM-as-a-Judge across general performance, bias quantification, challenging tasks, and domain-specific settings).
However, these surveys focus mostly on English without systematically addressing LLM-as-a-Judge in multilingual or low-resource settings.

Alongside these surveys, a number of \textbf{empirical studies} and position papers have raised important caveats about the paradigm's reliability. \citet{bavaresco-etal-2025-llms} conduct a large-scale empirical study across 20 NLP tasks and find that LLM judges show substantial variability across tasks and datasets, cautioning that they should be carefully validated before deployment. 
\citet{chehbouni2025neither} argue that LLM-as-a-Judge has been widely adopted before its validity and reliability as an evaluation method have been thoroughly examined.

Although recent benchmarks such as MM-Eval \citep{son2024mm} quantify cross-lingual judge reliability, no prior study has surveyed the research landscape of LLM-as-a-Judge specifically in multilingual settings and for low-resource languages.

\section{Literature Search and Annotation Methodology}\label{sec:search-annotation}
\paragraph{Definition} \textbf{LLM-as-a-Judge} is used broadly to cover both evaluator-oriented and annotator-oriented uses of LLMs in the literature. 
\textit{Evaluator}-oriented settings use LLM judgments to measure, compare, or validate items (e.g., model responses, system outputs, retrieved evidence, or benchmark examples). 
Given task-specific context (e.g., an instruction, source text, candidate response, reference answer, rubric, or label definitions), the judge produces an evaluative output (e.g., a score, label, ranking, preference judgment, or textual assessment).
\textit{Annotator}-oriented settings use LLMs to produce labels, metadata, explanations, or error annotations for downstream analysis or training.
In this paper, we focus on the first meaning and only analyze LLM-as-evaluator settings because evaluator-oriented uses introduce an additional layer of meta-evaluation. The LLM is used to assess outputs produced by another system or process, and the  reliability of the judge directly affects conclusions concerning that system. Therefore, we focus on the validity and reliability of LLMs in the evaluator role, while annotation-oriented use cases are closer to standard labeling or classification settings.

\paragraph{Literature Search Methodology} 
We conducted a keyword-based search over metadata from the ACL Anthology \cite{bollmann-etal-2023-two}, which provides comprehensive and structured coverage of research in NLP, including work on multilinguality and low-resource languages. Our goal is not to exhaustively enumerate all relevant publications across venues, but to obtain a representative overview of LLM-based evaluation research within the NLP community. We implemented the search as a lightweight, fully reproducible pipeline in our project repository,%
\footnote{\url{https://github.com/verenablaschke/aclanthology-llmasajudge}} 
which contains the code for parsing Anthology XML files and performing keyword matching. 

\paragraph{Data Source and Reproducibility} Our search operates on the official ACL Anthology repository (\href{https://github.com/acl-org/acl-anthology/blob/master/LICENSE}{Apache~2.0 license}), which provides structured XML metadata (e.g., titles, abstracts, and venue information) for papers published at major ACL venues. To ensure reproducibility, we fixed the Anthology snapshot to one commit (\href{https://github.com/acl-org/acl-anthology/tree/370911ef38556af764c17d456be9fce2d477b0bd}{370911e}, 2025-11-14). %
It includes the major ACL conferences (e.g., ACL, EMNLP, NAACL, EACL, AACL), two peer-reviewed journals (\textit{Computational Linguistics} and \textit{TACL}), other recurring NLP conferences (e.g., LREC, COLING), and several hundred specialized workshops.

\paragraph{Keyword Design and Matching} The search is applied to the concatenation of each paper's title and abstract, and is driven by three manually curated keyword groups targeting (i) LLMs, (ii) evaluation or judging functionality, and (iii) low-resource or multilingual contexts. 
Matching is case-insensitive, and a paper is counted as a hit if it contains at least one keyword from each of the three groups: 

\begin{itemize}
    \item \textbf{LLM}: \texttt{[``LLM'', ``large language model'']}
    \item \textbf{Judge}: \texttt{[``judge'', ``evaluator'', ``LLM-based evaluation'', ``LLM-as-a-judge'', ``LLM-based assessment'']}
    \raggedright
    \item \textbf{Low-resource}:\\ \texttt{[``low-resource'',  ``low resource'', ``underresourced'',  ``under-resourced'', ``underresearched'', ``under-researched'', ``multilingual'']}
\end{itemize}

\noindent
We include \textit{multilingual} in the low-resource group to increase recall, as other low-resource-specific terms alone produce relatively few matches. We also exclude some candidate keywords (e.g., \textit{annotator}) that frequently triggered false positives, since they typically refer to human annotators rather than to automated or LLM-based evaluators.

Enforcing low-resource related keywords substantially reduces the number of retrieved papers. 
Specifically, across the full ACL Anthology directory, the number of hits decreases from 650 to 49 under the same constraint.\footnote{The raw search returned 50 results, but one paper (\citealt{song-etal-2025-mug-eval}) appeared twice (once in the Findings of ACL and once in the Proceedings of the Workshop on Multilingual Representation Learning). Therefore, we counted it only once, yielding 49 unique papers.}
This indicates that comparatively few papers explicitly position LLM-based evaluation methods in low-resource or underrepresented language settings.

\label{sec:annotation}
\paragraph{Annotation and Exclusion Criteria} We manually reviewed each of the 49 candidate papers and annotated the role of the LLM, the task being judged, the languages covered, and the validation protocol against human judgments or gold-standard benchmark labels. We identified 33 papers in which an LLM is used to assess model-generated or human-produced outputs. The remaining 16 papers were excluded for the following reasons.

We excluded annotation-oriented papers in which LLMs are used to directly label text data rather than evaluate outputs of a previous processing step. 
These include work using LLMs for toxicity ratings \citep{bell-etal-2025-translate, faisal-etal-2025-dialectal}, multiclass classification \citep{upadhayay-behzadan-2025-x}, and binary decisions \citep{tran-nam-2025-l3i}.

Some papers contain both the concepts of ``LLMs'' and ``judgment,'' but they do not necessarily investigate ``LLM-as-a-Judge''.  For example, several papers examine human judgments of LLM output quality \citep{anh-etal-2024-morphology,thakur-etal-2024-knowing}, moral judgments made by LLMs \citep{agarwal-etal-2024-ethical}, LLMs' judgments of semantic similarity \citep{brglez-etal-2024-human,wang-etal-2024-rethinking}, or grammaticality and acceptability \citep{zhang-etal-2024-mela}. 
However, they do not use LLMs as evaluators of model outputs. Therefore, they fall outside our definition of LLM-as-a-Judge.

We also excluded papers that do not focus on natural language (e.g., evaluating programming languages \citep{chen-etal-2024-jumpcoder} or domain-specific query languages \citep{li-etal-2025-hallucination}) and  two papers\footnote{The string ``LLM'' appeared only as part of the cited name ``Fi\textbf{llm}ore'' \citep{pimentel-2012-identifying,minnema-etal-2022-sociofillmore}.} that did not use LLMs.

Finally, two papers explicitly avoid LLM-as-a-Judge evaluation in favor of objective evaluation, citing self-enhancement bias \citep{dussolle-etal-2025-ifeval} and low reliability in low-resource languages \citep{song-etal-2025-mug-eval} as reasons.

\definecolor{rootcol}{RGB}{67,76,144}      %
\definecolor{L1col}{RGB}{118,120,180}      %
\definecolor{L2col}{RGB}{189,178,219}      %
\definecolor{leafcol}{RGB}{248,245,252}    %
\definecolor{leafborder}{RGB}{189,178,219} %
\definecolor{edgecol}{RGB}{140,140,160}    %

\definecolor{pastelblue}{RGB}{118, 149, 180}
\definecolor{whiteblue}{RGB}{237, 245, 252}
\definecolor{pastelgreen}{RGB}{118, 180, 137}
\definecolor{whitegreen}{RGB}{237, 255, 243}
\definecolor{pastelyellow}{RGB}{180, 177, 118}
\definecolor{whiteyellow}{RGB}{252, 252, 237}
\definecolor{pastelred}{RGB}{180, 140, 118}
\definecolor{whitered}{RGB}{255, 247, 242}

\begin{figure*}[t!]
\centering
\resizebox{0.97\textwidth}{!}{%
\begin{forest}
  for tree={
    grow'=0,
    child anchor=north,
    parent anchor=south,
    edge path={
      \noexpand\path [draw, edgecol, line width=0.6pt, rounded corners=4pt]
        (!u.parent anchor) -- +(8pt,0) |- (.child anchor)\forestoption{edge label};
    },
    l sep=14pt,
    s sep=4pt,
    inner xsep=6pt,
    inner ysep=4pt,
    font=\footnotesize,
  },
  root/.style={
    fill=rootcol, text=white, rounded corners=4pt,
    font=\normalsize\bfseries, 
    align=center, calign=center, inner ysep=6pt,
    rotate=90,
  },
  L1/.style={
    text=white, rounded corners=3pt,
    font=\normalsize\bfseries, 
    align=center, calign=center, inner ysep=4pt,
    rotate=90,
  },
  L2/.style={
    draw=leafborder,
    rounded corners=3pt,
    font=\small, text width=16.7cm,
    align=left, inner ysep=3pt,
    child anchor=west,
  },
  [{Multilingual LLM-as-a-Judge}, root, calign=center
    [{Judging paradigm}, fill=L1col, L1
      [{\textbf{\normalsize Direct scoring.} \citet{boughorbel-hawasly-2023-analyzing,
        boughorbel-etal-2024-improving, dinh-etal-2024-sciex, gupta-etal-2024-walledeval}; \citeauthor{hada-etal-2024-large} \\ 
        (\citeyear{hada-etal-2024-large});
        \citet{mendonca-etal-2024-ecoh, qian-etal-2024-large, sato-etal-2024-tmu, devine-2024-sure, watts-etal-2024-pariksha}; \\
        \citet{cruz-blandon-etal-2025-memerag, fu-liu-2025-reliable, kocmi-etal-2025-findings-wmt25}; \\
        \citet{niculae-etal-2025-dr, niklaus-etal-2025-swiltra, ellinger-groh-2025-depends, farhan-2025-hyderabadi}}, fill=leafcol, L2
      ]
      [{\textbf{\normalsize Multi-dimensional judging.} \citet{boughorbel-etal-2024-improving, hada-etal-2024-metal, liu-etal-2024-optimizing, cruz-blandon-etal-2025-memerag}; \\
        \citet{doddapaneni-etal-2025-cross, duwal-etal-2025-domain, kocmi-etal-2025-findings-wmt25, marquez-etal-2025-nlp}; \\
        \citet{mukherjee-etal-2025-evaluating, niklaus-etal-2025-swiltra, okewunmi-etal-2025-evaluating, rolshoven-etal-2025-unlocking, sitaram-etal-2025-multilingual}; \\
        \citet{umutlu-etal-2025-evaluating}}, fill=leafcol, L2
      ]
      [{\textbf{\normalsize Pairwise comparison and ranking.} \citet{devine-2024-sure, forde-etal-2024-evaluating, marchisio-etal-2024-quantization, raju-etal-2024-constructing}; \\
        \citet{watts-etal-2024-pariksha, marquez-etal-2025-nlp, thakur-etal-2025-mirage, bennie-etal-2025-codeofconduct, sitaram-etal-2025-multilingual}}, fill=leafcol, L2
      ]
      [{\textbf{\normalsize Multi-step/reasoning-based.} \citet{raju-etal-2024-constructing, cruz-blandon-etal-2025-memerag, lu-etal-2025-mqm, niklaus-etal-2025-swiltra}}, fill=leafcol, L2
      ]
      [{\textbf{\normalsize Multi-agent/ensemble.} \citet{niklaus-etal-2025-swiltra}}, fill=leafcol, L2
      ]
    ]
    [{Downstream NLP tasks}, fill=pastelblue, L1
      [{\textbf{\normalsize Machine translation.} \citet{marchisio-etal-2024-quantization, qian-etal-2024-large, sato-etal-2024-tmu}; \\
        \citet{fu-liu-2025-reliable, kocmi-etal-2025-findings-wmt25, lu-etal-2025-mqm, niklaus-etal-2025-swiltra}}, fill=whiteblue, L2
      ]
      [{\textbf{\normalsize Summarization.} \citet{forde-etal-2024-evaluating, hada-etal-2024-large, hada-etal-2024-metal,fu-liu-2025-reliable, kocmi-etal-2025-findings-wmt25, rolshoven-etal-2025-unlocking}; \\
        \citet{sitaram-etal-2025-multilingual, umutlu-etal-2025-evaluating}}, fill=whiteblue, L2
      ]      
      [{\textbf{\normalsize Safety/toxicity.} \citet{gupta-etal-2024-walledeval,  bennie-etal-2025-codeofconduct, farhan-2025-hyderabadi, marquez-etal-2025-nlp}}, fill=whiteblue, L2
      ]     
      [{\textbf{\normalsize Instruction following.} \citet{boughorbel-hawasly-2023-analyzing, boughorbel-etal-2024-improving,  devine-2024-sure, hada-etal-2024-large}; \\ 
        \citet{mendonca-etal-2024-ecoh, raju-etal-2024-constructing, doddapaneni-etal-2025-cross, duwal-etal-2025-domain, fu-liu-2025-reliable}; \\
        \citet{kocmi-etal-2025-findings-wmt25}}, fill=whiteblue, L2
      ] 
      [{\textbf{\normalsize QA \& reasoning.}\ \citet{dinh-etal-2024-sciex, marchisio-etal-2024-quantization, watts-etal-2024-pariksha, fu-liu-2025-reliable, okewunmi-etal-2025-evaluating}}, fill=whiteblue, L2
      ]
      [{\textbf{\normalsize RAG.} \citet{cruz-blandon-etal-2025-memerag, thakur-etal-2025-mirage}}, fill=whiteblue, L2
      ]
      [{\textbf{\normalsize Other.} Style transfer \citep{mukherjee-etal-2025-evaluating}, referential ambiguity resolution \citep{ellinger-groh-2025-depends}, \\ L2 tutoring \citep{liu-etal-2024-optimizing}, benchmark / dataset quality evaluation \citep{umutlu-etal-2025-evaluating}}, fill=whiteblue, L2
      ]
    ]
    [{Judge LLMs}, fill=pastelgreen, L1
      [{\textbf{\normalsize Closed-source.} GPT-3.5/4.x/4o/4v (\citealp{boughorbel-hawasly-2023-analyzing, boughorbel-etal-2024-improving, devine-2024-sure, dinh-etal-2024-sciex}; \\
        \citealp{forde-etal-2024-evaluating, gupta-etal-2024-walledeval, hada-etal-2024-large, hada-etal-2024-metal, liu-etal-2024-optimizing, marchisio-etal-2024-quantization, mendonca-etal-2024-ecoh}; \\
        \citealp{raju-etal-2024-constructing, sato-etal-2024-tmu,watts-etal-2024-pariksha, cruz-blandon-etal-2025-memerag, doddapaneni-etal-2025-cross, duwal-etal-2025-domain}; \\
        \citealp{ellinger-groh-2025-depends, fu-liu-2025-reliable, kocmi-etal-2025-findings-wmt25, mukherjee-etal-2025-evaluating, niklaus-etal-2025-swiltra}; \\
        \citealp{okewunmi-etal-2025-evaluating, sitaram-etal-2025-multilingual, thakur-etal-2025-mirage, umutlu-etal-2025-evaluating}), 
        Gemini (\citealp{doddapaneni-etal-2025-cross}; \\
        \citealp{fu-liu-2025-reliable, niklaus-etal-2025-swiltra, okewunmi-etal-2025-evaluating}),
        Claude (\citealp{gupta-etal-2024-walledeval, kocmi-etal-2025-findings-wmt25}; \\          
        \citealp{okewunmi-etal-2025-evaluating}),
        PaLM2 (\citealp{hada-etal-2024-metal}),
        Mistral Medium (\citealp{kocmi-etal-2025-findings-wmt25})}, fill=whitegreen, L2
      ]
      [{\textbf{\normalsize Open-source.} Llama-2/3.x/4 (\citealp{dinh-etal-2024-sciex, qian-etal-2024-large, raju-etal-2024-constructing, cruz-blandon-etal-2025-memerag}; \\ 
        \citealp{doddapaneni-etal-2025-cross, fu-liu-2025-reliable, kocmi-etal-2025-findings-wmt25, lu-etal-2025-mqm, liu-etal-2024-optimizing}; \citeauthor{mukherjee-etal-2025-evaluating} \\
        (\citeyear{mukherjee-etal-2025-evaluating}); \citealp{thakur-etal-2025-mirage, umutlu-etal-2025-evaluating}),
        Qwen 1.5/2.5/3 (\citealp{mendonca-etal-2024-ecoh, qian-etal-2024-large}; \citeauthor{cruz-blandon-etal-2025-memerag} \\
        (\citeyear{cruz-blandon-etal-2025-memerag}); \citealp{fu-liu-2025-reliable, kocmi-etal-2025-findings-wmt25, lu-etal-2025-mqm, sitaram-etal-2025-multilingual}), Mistral 7B/Mixtral (\citealp{dinh-etal-2024-sciex}; \\
        \citealp{qian-etal-2024-large, kocmi-etal-2025-findings-wmt25, lu-etal-2025-mqm, liu-etal-2024-optimizing}), 
        DeepSeek v3 (\citealp{kocmi-etal-2025-findings-wmt25}; 
        \citeauthor{rolshoven-etal-2025-unlocking},\\ \citeyear{rolshoven-etal-2025-unlocking}), 
        Gemma (\citealp{niculae-etal-2025-dr, qian-etal-2024-large, umutlu-etal-2025-evaluating}),
        Command/AyaExpanse
        (\citealp{fu-liu-2025-reliable};\\ \citealp{kocmi-etal-2025-findings-wmt25}), 
        TowerInstruct-7B/13B (\citealp{lu-etal-2025-mqm}),
        OpenChat 3.5 (\citealp{qian-etal-2024-large}),
        Phi \cite{sitaram-etal-2025-multilingual}}, 
        fill=whitegreen, L2
      ]
    [{\textbf{\normalsize Specialized/Trained-as-judge.} Llama Guard (Llama-based) / WalledGuard (Qwen-based)  \citep{gupta-etal-2024-walledeval}, JudgeLM \\
      (Vicuna-based) \citep{bennie-etal-2025-codeofconduct, farhan-2025-hyderabadi, marquez-etal-2025-nlp},  MedGemma (Gemma-based) (\citeauthor{niculae-etal-2025-dr}, \\
      \citeyear{niculae-etal-2025-dr}),
      ECoh (Qwen-based) \citep{mendonca-etal-2024-ecoh}, 
      Hercule and Prometheus (Llama-based) \citep{doddapaneni-etal-2025-cross}}, fill=whitegreen, L2
    ] 
    ]
    [{Language Coverage}, fill=pastelyellow, L1
      [{\textbf{\normalsize Monolingual.} \citet{boughorbel-etal-2024-improving, duwal-etal-2025-domain, niculae-etal-2025-dr, umutlu-etal-2025-evaluating}}, fill=whiteyellow, L2
      ]
      [{\textbf{\normalsize Moderately multilingual} (2--10 languages). \citet{boughorbel-hawasly-2023-analyzing, devine-2024-sure, dinh-etal-2024-sciex}; \\
        \citet{forde-etal-2024-evaluating, gupta-etal-2024-walledeval, hada-etal-2024-metal, hada-etal-2024-large, liu-etal-2024-optimizing, mendonca-etal-2024-ecoh, qian-etal-2024-large}; \\
        \citet{raju-etal-2024-constructing, sato-etal-2024-tmu, watts-etal-2024-pariksha, bennie-etal-2025-codeofconduct, cruz-blandon-etal-2025-memerag}; \\
        \citet{doddapaneni-etal-2025-cross, ellinger-groh-2025-depends, farhan-2025-hyderabadi, lu-etal-2025-mqm, marquez-etal-2025-nlp}; \\
        \citet{mukherjee-etal-2025-evaluating, niklaus-etal-2025-swiltra, okewunmi-etal-2025-evaluating, rolshoven-etal-2025-unlocking}}, fill=whiteyellow, L2
      ]
      [{\textbf{\normalsize Massively multilingual} (10+ languages). \citet{marchisio-etal-2024-quantization, devine-2024-sure, fu-liu-2025-reliable, kocmi-etal-2025-findings-wmt25}; \\
        \citet{sitaram-etal-2025-multilingual, thakur-etal-2025-mirage}}, fill=whiteyellow, L2
      ]
    ]
    [{Evaluation of the Judge}, fill=pastelred, L1
      [{\textbf{\normalsize Agreement with human/gold labels} (correlation or agreement metrics with human/gold-label references for at least one\\
        language). \citet{forde-etal-2024-evaluating, hada-etal-2024-large, hada-etal-2024-metal, marchisio-etal-2024-quantization, raju-etal-2024-constructing, dinh-etal-2024-sciex}; \citeauthor{gupta-etal-2024-walledeval} \\
        (\citeyear{gupta-etal-2024-walledeval}); \citet{liu-etal-2024-optimizing, mendonca-etal-2024-ecoh, qian-etal-2024-large, sato-etal-2024-tmu, watts-etal-2024-pariksha}; \citeauthor{cruz-blandon-etal-2025-memerag} \\
        (\citeyear{cruz-blandon-etal-2025-memerag}); \citet{doddapaneni-etal-2025-cross, ellinger-groh-2025-depends, kocmi-etal-2025-findings-wmt25, lu-etal-2025-mqm, mukherjee-etal-2025-evaluating}; \\
        \citet{niculae-etal-2025-dr, niklaus-etal-2025-swiltra, rolshoven-etal-2025-unlocking}; 
        \citet{sitaram-etal-2025-multilingual, umutlu-etal-2025-evaluating}}, fill=whitered, L2
      ]
      [{\textbf{\normalsize No direct evaluation} 
      (Not assessed against any reference).
      \citet{boughorbel-hawasly-2023-analyzing}; \citeauthor{boughorbel-etal-2024-improving}
      (\citeyear{boughorbel-etal-2024-improving}); \\ 
      \citet{bennie-etal-2025-codeofconduct, duwal-etal-2025-domain, farhan-2025-hyderabadi, marquez-etal-2025-nlp, okewunmi-etal-2025-evaluating}; \citeauthor{thakur-etal-2025-mirage} \\
      (\citeyear{thakur-etal-2025-mirage})
      }, fill=whitered, L2
      ]
      [{\textbf{\normalsize 
      Internal consistency/robustness} 
      (Self-consistency across rankings/languages/prompts).
      \citet{devine-2024-sure, fu-liu-2025-reliable}}, fill=whitered, L2
      ]
    ]
  ]
\end{forest}
}%

\caption{Taxonomy of the 33 papers that use LLM-as-a-Judge in an \textbf{evaluator} role (i.e., assessing model-generated or human-produced outputs) in multilingual settings. Papers are organized along five dimensions:
  \textbf{Judging Paradigm} (how the judge scores),
  \textbf{Downstream NLP Tasks},
  \textbf{Judge LLMs} (model families used),
  \textbf{Language Coverage}, and
  \textbf{Evaluation of the Judge} (how judge reliability is assessed).
  A paper may appear in multiple categories.
 }
\label{fig:taxonomy}
\end{figure*}
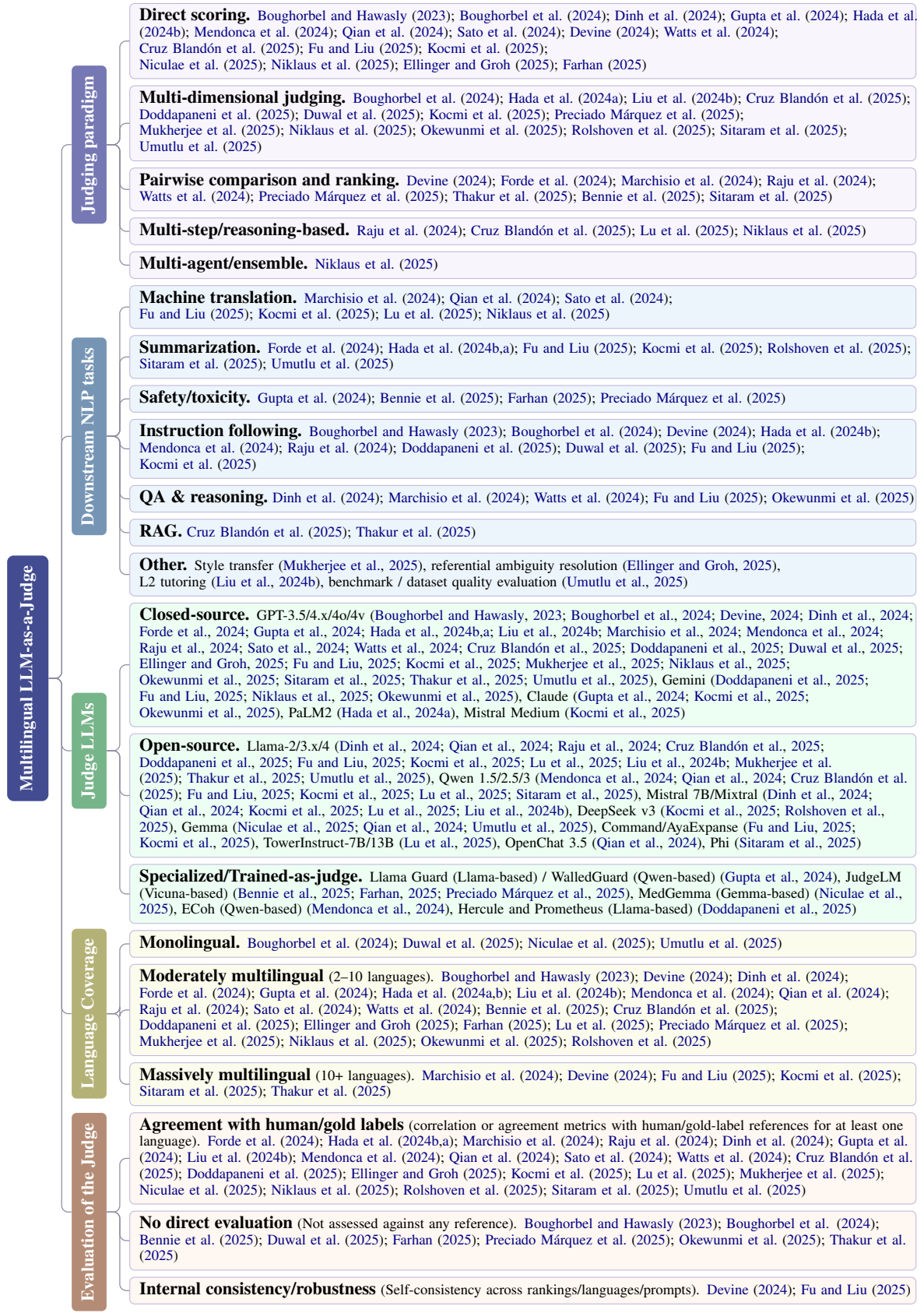

\section{LLM-as-a-Judge in Multilingual Research}\label{sec:findings}

We analyze the 33 in-scope papers along five dimensions (i.e., judging paradigm, downstream task, judge model, language coverage, and judge evaluation).
This section organizes our findings around three cross-cutting claims that draw on these dimensions and motivate the recommendations in Section~\ref{sec:discussion}.
Figure~\ref{fig:taxonomy} provides an extensive  distribution of the papers across these five dimensions, with a breakdown of language coverage in Appendix~\ref{app:lr-coverage} and detailed annotations for each paper in Appendix~\ref{app:annotations}.
 
We focus on five judging paradigms: \textbf{direct scoring} (the LLM judge produces a scalar score, label, or short verdict from input and candidate output, sometimes with a reference or rubric as in  \citealt{dinh-etal-2024-sciex, gupta-etal-2024-walledeval, qian-etal-2024-large}), \textbf{pairwise comparison and ranking} (two or more candidate outputs are compared, sometimes with ties as in \citealt{raju-etal-2024-constructing, watts-etal-2024-pariksha, thakur-etal-2025-mirage}), \textbf{multi-step or reasoning-based judging} (the judge generates explicit reasoning steps before issuing a verdict as in \citealt{lu-etal-2025-mqm}), \textbf{multi-dimensional judging} (the judge rates several predefined quality aspects separately rather than evaluating quality holistically in one step as in \citealt{boughorbel-etal-2024-improving, mukherjee-etal-2025-evaluating, okewunmi-etal-2025-evaluating}), and \textbf{multi-agent or ensemble judging} (verdicts from multiple distinct judges are aggregated into a single final label as in \citealt{niklaus-etal-2025-swiltra}). 

The downstream tasks evaluated by these judges cover tasks such as machine translation (MT), summarization, instruction following, question answering, retrieval-augmented generation, safety and moderation as well as more specialized tasks (e.g., text style transfer, referential ambiguity resolution, tutoring, and benchmark quality assessment). The distribution of papers and tasks is provided in Figure~\ref{fig:taxonomy}. 

The models used as judges include both generalist instruction-tuned LLMs and more specialized judge models. Generalist judges include closed-source models (e.g., GPT, Gemini, and Claude), as well as open-source models from families (e.g., Llama and Qwen). 
Specialized judges (e.g.,Llama Guard for safety evaluation and domain-adapted evaluators for specialized tasks) are used in narrower settings where the output space is more constrained.

Based on our analyses, LLM-as-a-Judge is already used broadly across NLP tasks, different judge model families, and diverse evaluation settings.
However, a closer reading of the surveyed papers reveals that this breadth also masks several structural weaknesses which we report below.

\paragraph{Broad Multilingual Coverage but Shallow Depth for Low-Resource Languages} 
Broad multilingual coverage in the surveyed literature does not necessarily indicate a focus on low-resource languages.
We classify languages using \citet{joshi-etal-2020-state}'s taxonomy, treating its Class 0–3 as low-resource.\footnote{We use a binary classification of language resource level, assigning each language to either the low-resource or high-resource category. While this pre-LLM classification, based on labeled-data availability, may not fully capture resourcedness of languages at the time of writing, we still adopt it because it remains the most widely cited reference in the literature and provides a transparent, reproducible criterion for our purposes. 
}
Among the 19/33 papers (58\%) that include at least one low-resource language, coverage is still often skewed toward higher-resource ones. For example,  
\citet{sitaram-etal-2025-multilingual} cover 37 languages, 
\citet{kocmi-etal-2025-findings-wmt25} 30,
and \citet{fu-liu-2025-reliable} 25.
However, only 36--47\% of these language lists can be considered as low-resource. Similarly, \citet{marchisio-etal-2024-quantization} cover 22 languages but only 4 (18\%) of them can be considered as low-resource. Only 8 of the 33 surveyed papers (24\%) include low-resource languages as at least half of their evaluation languages.
The distribution gets even more skewed under a stricter cutoff that excludes Joshi Class 3, the so-called ``Rising Stars'' of low-resource NLP (e.g.,  Indonesian, Thai), which already have substantial LLM support. Only 13/33 papers (39\%) include at least one low-resource language under this stricter definition.

The low-resource languages covered in the surveyed papers are themselves unevenly distributed. Among the 8 papers where low-resource languages represent at least half of the languages studied, 6 focus primarily on South Asian languages \citep{watts-etal-2024-pariksha, sato-etal-2024-tmu, qian-etal-2024-large, lu-etal-2025-mqm, doddapaneni-etal-2025-cross, duwal-etal-2025-domain}. The remaining cases represent smaller pockets of research on Yorùbá \citep{okewunmi-etal-2025-evaluating} and Romanian \citep{niculae-etal-2025-dr}.
Many low-resource languages are entirely or nearly entirely absent. Indigenous American and Pacific Island languages do not appear in any of the papers in the survey. Sub-Saharan African languages are represented only by Maasai, Swahili, and Yorùbá. These are precisely the settings where dominant judge models are the weakest and where the human reference data needed to validate them is scarcest \citep{adelani-etal-2024-comparing, smart2024socially, ojo-etal-2025-afrobench}.

\paragraph{Narrow and Closed LLMs in the Judge Ecosystem}
The current judge ecosystem is dominated by closed and proprietary models. More precisely, GPT-family models are used in 26/33 papers (79\%), and 13/33 (39\%) use \emph{only} closed-source judges, compared with just 3/33 (9\%) that use \emph{only} open-source ones.
Cross-judge validation is not a common practice. 15/33 papers (45\%) rely on a single judge model family, and 10/33 (30\%) use GPT as their sole judge.
Among open-source generalist alternatives, Llama (13/33) and Qwen (7/33) are the most common. However, in most papers they appear alongside GPT rather than on their own (i.e., 10 of the 13 Llama papers and 5 of the 7 Qwen papers also use a GPT model as judge).
The majority of judges are generalist instruction-tuned LLMs, repurposed via rubrics or comparison prompts. Only 7/33 papers (21\%) use a model that is fine-tuned or specialized for the evaluation task itself, including safety classifiers \citep{gupta-etal-2024-walledeval}, judge-tuned models like JudgeLM \citep{bennie-etal-2025-codeofconduct, farhan-2025-hyderabadi, marquez-etal-2025-nlp}, a domain-fine-tuned variant \citep{niculae-etal-2025-dr}, and evaluator models fine-tuned for specialized evaluation tasks (e.g., multilingual coherence \citep{mendonca-etal-2024-ecoh} or cross-lingual evaluation \citep{doddapaneni-etal-2025-cross}).

\paragraph{Problems with Validation} %
The third pattern concerns how judge reliability is checked. On the surface, validation appears to be widespread. That is to say, 23/33 papers (70\%) report some form of human/gold-label comparison or expert evaluation of the judge in at least one deployment language. 
However, this aggregate figure could be obscuring a subtler problem. Sometimes the more revealing question is not \emph{whether} papers validate, but \textbf{\emph{which} languages they validate on}.
Some papers that deploy LLM-as-a-Judge in multilingual settings check reliability only on a subset of languages. For example, \citet{mendonca-etal-2024-ecoh} build a dialogue coherence judge for five languages (i.e., English, French, German, Italian, and Chinese). However, the check against human ratings is performed on an existing English-only benchmark of human-annotated dialogue responses. In other words, the reliability of LLM-as-a-Judge in the other languages (i.e., French, German, Italian, and Chinese) is never compared to a human judgment. This choice is convenient since a human-annotated English benchmark is readily available.
However, English is also the language where the reliability of LLM-as-a-Judge is perhaps the least controversial (considering how extensively frontier models have been evaluated on it).
As a result, the validation only takes place where it is least needed. On the other hand, the languages for which an independent check of validity would be more informative are left unchecked.

The same pattern is also observed when LLM-as-a-Judge is deployed directly in low-resource languages. Of the eight papers that report no human or gold-label check at all, three do so while judging outputs in low-resource languages.
\citet{duwal-etal-2025-domain} use GPT-4o as the only judge for Nepali generation. Instead of comparing its Nepali outputs to human or gold labels, this evaluation choice is justified by citing prior work on GPT-4o's multilingual coverage.
\citet{thakur-etal-2025-mirage} deploy an LLM judge across 18 languages, including low-resource ones (e.g., Telugu, Swahili, Yorùbá, Bengali, Indonesian, and Thai). They treat the judge's pairwise preferences as gold labels by default, so the judge's reliability is assumed rather than tested.
\citet{okewunmi-etal-2025-evaluating} use GPT4.0, Gemini~2.0 Flash, and Claude~3.7 Sonnet to score Yorùbá question answering without human evaluation of the outputs by the LLM judges. 

These decisions are difficult to justify, because the existing evidence (both within our corpus and outside it) indicates that judge reliability is language-conditional rather than uniform. Among the papers that test judges directly in low-resource languages, this finding is consistent. \citet{watts-etal-2024-pariksha} report that human-LLM agreement drops for direct assessment, particularly for Bengali and Odia. \citet{fu-liu-2025-reliable} report an average Fleiss' Kappa of approximately 0.3 across 25 languages, with consistency being particularly poor in low-resource languages, and find that neither multilingual training nor model scale directly improves this result. Studying GPT-4 as an evaluator across nine languages, \citet{hada-etal-2024-large} document a bias toward higher scores when human opinions differ and it is most pronounced in lower-resource and non-Latin-script languages.

Therefore, the problem is not that validation is rare. It is rather the case that validation is often skipped for low-resource languages. In these cases, it is not clear that the LLM-as-a-Judge evaluation can be trusted without human validation (given the underrepresentation of low-resource languages in LLM training).   %

\section{Discussion and Recommendations}
\label{sec:discussion}

Evaluation of an NLP system is inherently difficult, no matter if it is carried out by humans or LLMs. To be able to perform the evaluation properly with humans takes a lot of time, requires effort and resources (e.g., recruiting and training qualified and representative/suitable annotators, collecting multiple annotations per items for a task in a language, identifying and understanding disagreements between annotators, making decisions about how the annotations are aggregated and used for evaluation). 
Therefore, it could be tempting to ignore many of these difficulties with human judgements and opt for LLMs which seem faster and cheaper. However, by reviewing the relevant literature (section~\ref{sec:search-annotation}), we have identified several recurring issues with this practice for languages other than English, particularly for low-resource ones. (section~\ref{sec:findings}).  %
Based on our survey, we provide the following recommendations for future research using LLM-as-a-Judge.

\paragraph{Recommendation 1:} \textit{Validate LLMs in multilingual contexts and for low-resource languages.}
While using LLM-as-a-Judge, it is crucial to investigate whether a judge is reliable for a given language.
For instance, \citet{zheng-etal-2023-judging} evaluate LLM-as-a-Judge on an English multi-turn question answering dataset.
\citet{boughorbel-hawasly-2023-analyzing} translate this dataset into Arabic and use the same LLM for judging model outputs. However, their claim about the reliability of a judge is only based on the English dataset and is not re-investigated for Arabic.
Given the fact that LLM judges tend to overestimate the quality of text generated by the same LLM \cite{liu-etal-2024-llms-narcissistic, panickssery2024llm}, it is especially important to properly validate a judge's performance on languages that are only covered by few language models, where the set of potential generator and judge models is extremely small.
In multilingual contexts, it is particularly important to compare the validity of LLM-as-a-Judge across the target languages \citep{mendonca-etal-2024-ecoh,cruz-blandon-etal-2025-memerag}.

\paragraph{Recommendation 2:} \textit{Keep humans in the loop.} %
Combining metrics and including human evaluations is a standard procedure in fields like speech synthesis, where automatic metrics capture different properties of the output and an ideal output depends on the context in which a system is deployed \cite{wagner-etal-2019-speech}. Similarly, shared tasks for MT systems often complement automatic evaluation metrics with human evaluation on a subset of the data to rank participants' systems (e.g., \citealp{kocmi-etal-2025-findings-wmt25}). When LLM-as-a-Judge is used for low-resource languages, human evaluation (at least on a small subset of the data) and statistical confidence estimates should also be conducted \citep[cf.][]{zheng-etal-2023-judging,niklaus-etal-2025-swiltra}.

It is also important to acknowledge that humans might disagree on how to judge a given text \cite{plank-2022-problem} or may introduce their own biases \citep{chen-etal-2024-humans}.
This can depend on many factors, including lack of familiarity with the specific dialect a text is written in \cite{keleg-etal-2024-estimating-level}. 
When using LLM-as-a-Judge, there is a need to document \textit{which} humans the judge LLM is compared to (e.g., which human population the human evaluators represent (\citealp{dogruoz-etal-2023-representativeness})), and which populations are modelled (or not modelled) by the LLM judges.
Researchers should document their selection criteria, and both the language competence and domain expertise of the human evaluators, as well as any other relevant criteria beyond their academic status. %

\paragraph{Recommendation 3:} \textit{Check if non-LLM-based metrics are equally (or more) feasible.}
While LLM-as-a-Judge is a new paradigm for evaluating LLM-generated outputs or human annotations, there is a need to carefully consider which existing metrics can actually be replaced by LLM judges, and which ones should still be used alongside them. This depends highly on the competence of the LLM on the target language and their agreement with human judgment. 
For instance, traditional metrics (e.g, Exact Match and F1-score) in question answering may prefer short answers while LLM generations are generally more verbose. LLM-as-a-Judge has been shown to provide better and more flexible extraction of answers than these reference-based metrics \cite{adlakha-etal-2024-evaluating}. However, this is only validated for English, and even in this setting models often do not stay faithful to provided references.
If references are available for a dataset, conventional if flawed metrics (e.g., chrF) should still be used alongside LLM judges to catch cases where the generated text is in the wrong language variety.

The weaker the LLM's competence in a language, the more errors and biases are likely to be introduced by LLM-as-a-Judge evaluation (e.g., rewarding longer answers \cite{zheng-etal-2023-judging, chen-etal-2024-humans}), preferring their own generations in direct comparison \cite{liu-etal-2024-llms-narcissistic, panickssery2024llm}, or relying on their own internal knowledge and ignoring explicit references \cite{lee2026judging}.
Therefore, LLM-as-a-Judge should only be used when no better or equally good existing metric is available. In addition, it should be used in conjunction and comparison with validated metrics (if possible). However, it is better to avoid it when there is a high risk of conflicting information between task-specific references and the LLM's learned representation.

\paragraph{Recommendation 4:} \textit{Consider real-world relevance and representativeness beyond language.}
The amount of available language data is not the only relevant dimension with respect to using LLM-as-a-Judge in low-resource settings. Failures on different dimensions should also be distinguished. For instance, linguistic competence and cultural competence can be two independent dimensions.
\citet{watts-etal-2024-pariksha} show that LLM evaluators agree less with humans on evaluating responses with culturally specific nuances. LLMs reproduce and amplify social and cultural stereotypes in their outputs \citep{mitchell-etal-2025-shades} and even introduce these (without being asked) in text-to-image generation, purely based on the language the prompt is written in \citep{holtermann-etal-2026-sos}. Besides investigating cultural competence, future work could also focus on low-resource domains (e.g., how well LLM-as-a-Judge works in medical or legal contexts across languages, cf.~\citealp{diekmann-etal-2025-llms,spiegel-etal-2025-adaption,rolshoven-etal-2025-unlocking}).

\paragraph{Checklist}
To help authors design and report on experiments with LLM judges (especially for languages other than English), we provide a checklist in Appendix~\ref{app:checklist}.
This checklist also serves to help reviewers evaluate such experiments.  

\section{Conclusion}

Evaluation is an important and necessary part of NLP pipelines, and it can be challenging in multilingual and low-resource settings. Currently, LLMs are widely used in the evaluation process across NLP tasks. Our goal in this paper was to explore the use of LLM-as-a-Judge paradigm for multilingual and low-resource settings in the literature. While application of LLM-as-a-Judge increasingly includes low-resource languages, our survey shows that this does not automatically lead to an inclusive application, as validation of LLM judges is still mostly done for English only. 
Although LLM judges may reduce the annotation costs and may deliver faster results, this benefit is limited when their judgments require additional verification or when the target language is weakly supported by the LLM judge. Therefore, human validation remains essential, especially for low-resource languages, where judge reliability cannot be assumed by default. Ideally, studies should validate LLM judges on a representative subset of target-language examples, report agreement with human or gold-label references, and clearly document the limitations of automatic judgments. 
If this is not done, the ease and ubiquity of LLM-as-a-Judge evaluation may become its own downfall. Although there is an increase in multilingual NLP evaluations, many of the included languages are still low-resource. Therefore, LLM performance may not be properly validated on them before the LLMs are used as judges on these same languages. This cycle will degrade the evaluation quality while superficially pretending to increase the representation breadth.

In our paper, we focus on LLM-as-evaluator settings, where the judge is explicitly applied to the output of another generative language model with the purpose of evaluating that model. There are also LLM-as-annotator settings, where labels are assigned to a preexisting text. These two paradigms grow increasingly indistinguishable. There is an increase in the online texts which are written or co-authored by LMs\footnote{\url{https://www.pewresearch.org/data-labs/2026/08/20/how-much-of-the-internet-is-written-with-ai/}, accessed August 24, 2026} and labels assigned by annotator LLMs are often used to create benchmarks and, ultimately, evaluate other models. Therefore, future work could also focus on analyzing and re-defining LLM-as-annotator in close connection with LLM-as-evaluator.

We hope that the results of our study  will raise awareness about the limitation of LLMs-as-a-Judge paradigm in critical evaluation contexts and our recommendations will serve as guidelines for NLP researchers who work on low-resource languages and multilingual settings.

\section*{Limitations}

Although we performed a thorough search of the ACL Anthology, we might have missed papers by not searching for specific language names or including other search terms (e.g.,  ``dialects'', ``varieties'').
Furthermore, our keyword search only matches papers whose titles and/or abstracts are in English. 
We only cover papers that appeared in the ACL Anthology (in order to point out current research trends in the ACL community), but other papers (or paper drafts) using LLM-as-a-Judge in multilingual settings and/or for low-resource languages might be in other research venues or on preprint servers. Although we did our best to analyse the papers in our survey carefully, some annotation decisions required judgement, as papers vary widely in how they describe their setups and the relevant information was not always stated explicitly. Therefore, occasional inconsistencies in annotations may remain.

Our paper consists of an analysis of already published papers rather than new experiments. This is partially due to the challenges (e.g., lack of data, lack of good existing metrics for NLG tasks in LRLs, difficulties with finding human experts for evaluating the LLM judges) of working with LRL data and partially because we aimed at identifying the shortcomings of current research trends  in this area and provide recommendations for improvement.

\section* {Ethical Considerations}
The annotators of the ACL Anthology papers are the authors of this paper, who carried out the annotation work as part of their jobs. We use data from the ACL Anthology in accordance with one of its intended uses: academic research about NLP research (cf.\ \citealp{bollmann-etal-2023-two}).

In parts of this paper, LLMs were used for helping with phrasing and/or correcting English (e.g., grammar, mistakes), and for assisting with the preparation of figures and tables. 

\section*{Acknowledgments}
This work has benefitted from the participation of A.\ Seza Doğruöz, David Adelani, Verena Blaschke, Jakob Prange and Xixian Liao in Dagstuhl Seminar 25301 ``Linguistics and Language Models: What Can They Learn from Each Other?''. We thank Schloss Dagstuhl – Leibniz Center for Informatics for providing an inspiring research environment. David Adelani acknowledges the support of Schmidt Sciences AI2050 program. 
Xixian Liao's research has been promoted and financed by the Government of Catalonia through the Aina project, and by the MLLM4TRA project (PID2024-158157OB-C32), funded by MICIU/AEI/10.13039/501100011033/FEDER, UE.
Verena Blaschke's research was supported by European Research Council (ERC) Consolidator Grant DIALECT 101043235.

\bibliography{custom}

\appendix

\section{Per-paper Language Coverage}
\label{app:lr-coverage}

This appendix provides a per-paper breakdown of language coverage across the 33 in-scope papers (Table~\ref{tab:paper-language-coverage}).
For each paper, we list the languages covered, sorted by \citeposs{joshi-etal-2020-state} language resource taxonomy.
We treat languages of class 0--3 as low-resource.
Note that this taxonomy is from 2020 and might thus underestimate the current availability of language resources.
Also note that the framing of a language's resourcedness may be different in the papers surveyed.
E.g., the surveyed papers that include Basque (class~4) frame it as a low-resource language \cite{bennie-etal-2025-codeofconduct, farhan-2025-hyderabadi, marquez-etal-2025-nlp}.

\begin{table*}[h]
\centering
\small
\begin{tabular}{@{}lcp{10.5cm}@{}}
\toprule
\textbf{Paper} & \textbf{\#LR} & \textbf{Language resource classes \cite{joshi-etal-2020-state} and languages} \\
\midrule

\citet{sitaram-etal-2025-multilingual} & 15 & \textit{0:~Slovene~/ 1:~Norwegian, Welsh~/ 3:~Bulgarian, Danish, Estonian, Greek, Hebrew, Indonesian, Latvian, Lithuanian, Romanian, Slovak, Thai, Ukrainian~/} 4:~Catalan, Croatian, Czech, Dutch, Finnish, Hungarian, Italian, Korean, Polish, Portuguese, Russian, Serbian, Swedish, Turkish, Vietnamese~/ 5:~Arabic, Chinese, English, French, German, Japanese, Spanish \\
\citet{kocmi-etal-2025-findings-wmt25} & 14 & \textit{0:~Maasai~/ 1:~Bhojpuri, Kannada~/ 2:~Icelandic, Marathi~/ 3:~Bengali, Egyptian Arabic, Estonian, Greek, Indonesian, Lithuanian, Romanian, Thai, Ukrainian~/} 4:~Czech, Hindi, Italian, Korean, Persian, Russian, Serbian, Swedish, Turkish, Vietnamese~/ 5:~Chinese, English, French, German, Japanese, Spanish \\
\citet{watts-etal-2024-pariksha} & 9 & \textit{1:~Gujarati, Kannada, Malayalam, Odia, Telugu~/ 2:~Marathi, Punjabi~/ 3:~Bengali, Tamil~/} 4:~Hindi \\
\citet{fu-liu-2025-reliable} & 9 & \textit{1:~Telugu~/ 2:~Swahili~/ 3:~Bengali, Greek, Hebrew, Indonesian, Romanian, Thai, Ukrainian~/} 4:~Czech, Dutch, Hindi, Italian, Korean, Portuguese, Russian, Turkish, Vietnamese~/ 5:~Arabic, Chinese, English, French, German, Japanese, Spanish \\
\citet{thakur-etal-2025-mirage} & 6 & \textit{1:~Telugu~/ 2:~Swahili, Yoruba~/ 3:~Bengali, Indonesian, Thai~/} 4:~Finnish, Hindi, Korean, Persian, Russian~/ 5:~Arabic, Chinese, English, French, German, Japanese, Spanish \\
\citet{qian-etal-2024-large} & 5 & \textit{0:~Sinhala~/ 1:~Nepali~/ 2:~Marathi~/ 3:~Estonian, Romanian~/} 4:~Russian~/ 5:~Chinese, English, German \\
\citet{lu-etal-2025-mqm} & 4 & \textit{1:~Assamese, Kannada, Maithili~/ 2:~Punjabi~/} 4:~Russian~/ 5:~Chinese, English, German \\
\citet{marchisio-etal-2024-quantization} & 4 & \textit{3:~Greek, Indonesian, Romanian, Ukrainian~/} 4:~Czech, Dutch, Hindi, Italian, Korean, Persian, Polish, Portuguese, Russian, Turkish, Vietnamese~/ 5:~Arabic, Chinese, English, French, German, Japanese, Spanish \\
\citet{sato-etal-2024-tmu} & 3 & \textit{1:~Gujarati, Telugu~/ 3:~Tamil~/} 4:~Hindi~/ 5:~English \\
\citet{doddapaneni-etal-2025-cross} & 3 & \textit{1:~Telugu~/ 3:~Bengali, Urdu~/} 4:~Hindi~/ 5:~French, German \\
\citet{hada-etal-2024-metal} & 2 & \textit{2:~Swahili~/ 3:~Bengali~/} 4:~Hindi, Russian, Turkish~/ 5:~Arabic, Chinese, English, French, Japanese \\
\citet{raju-etal-2024-constructing} & 2 & \textit{3:~Slovenian, Thai~/} 4:~Hungarian, Russian, Serbian, Turkish~/ 5:~Arabic, Japanese \\
\citet{duwal-etal-2025-domain} & 1 & \textit{1:~Nepali} \\
\citet{niklaus-etal-2025-swiltra} & 1 & \textit{1: Romansh~/} 4:~Italian~/ 5:~English, French, German \\
\citet{okewunmi-etal-2025-evaluating} & 1 & \textit{2:~Yorùbá~/} 5:~English \\
\citet{forde-etal-2024-evaluating} & 1 & \textit{3:~Indonesian~/} 5:~Chinese, English  \\
\citet{gupta-etal-2024-walledeval} & 1 & \textit{3:~Filipino~/} 4:~Hindi, Russian, Serbian~/ 5:~Arabic, English, French, Spanish \\
\citet{mukherjee-etal-2025-evaluating} & 1 & \textit{3:~Bengali~/} 4:~Hindi~/ 5:~English \\
\citet{niculae-etal-2025-dr} & 1 & \textit{3:~Romanian} \\
\citet{devine-2024-sure} & 0 & 4:~Russian~/ 5:~Chinese, English, French, German, Japanese; also 62 non-specified languages \\
\citet{hada-etal-2024-large} & 0 & 4:~Czech, Italian, Portuguese~/ 5:~Chinese, English, French, German, Japanese, Spanish\\
\citet{liu-etal-2024-optimizing} & 0 & 4:~Korean~/ 5:~Chinese, English \\
\citet{mendonca-etal-2024-ecoh} & 0 & 4:~Italian~/ 5:~Chinese, English, French, German\\
\citet{bennie-etal-2025-codeofconduct} & 0 & 4:~Basque, Italian~/ 5:~English, Spanish~/  \\
\citet{cruz-blandon-etal-2025-memerag} & 0 & 4:~Hindi~/ 5:~English, French, German, Spanish \\
\citet{ellinger-groh-2025-depends} & 0 & 4:~Russian~/ 5:~Arabic, Chinese, English, French \\
\citet{farhan-2025-hyderabadi} & 0 &4:~Basque, Italian~/ 5:~English, Spanish \\
\citet{marquez-etal-2025-nlp} & 0 & 4:~Basque, Italian~/ 5:~English, Spanish \\
\citet{rolshoven-etal-2025-unlocking} & 0 & 4:~Italian~/ 5:~French, German \\
\citet{umutlu-etal-2025-evaluating} & 0 & 4:~Turkish \\
\citet{boughorbel-hawasly-2023-analyzing} & 0 & 5:~Arabic, English \\
\citet{boughorbel-etal-2024-improving} & 0 & 5:~Arabic \\
\citet{dinh-etal-2024-sciex} & 0 & 5:~English, German \\
\bottomrule
\end{tabular}
\caption{Language coverage per surveyed paper (n=33). Languages are sorted using \citeposs{joshi-etal-2020-state} taxonomy (0 = least resources, 5 = most resources). ``LR'' = low-resource, counting languages from classes 0--3 (language names in italics). Papers are sorted by number of LR languages covered (descending), then by lowest resource class included, then publication year (ascending), then alphabetically by author. For MT, both source and target languages are counted.}
\label{tab:paper-language-coverage}
\end{table*}

\section{Detailed Annotations}
\label{app:annotations}

Tables~\ref{tab:ann-setup} and \ref{tab:ann-reliability} give the full per-paper annotation underlying Figure~\ref{fig:taxonomy} and Section~\ref{sec:findings}, for all $n=33$ in-scope papers.
Table~\ref{tab:ann-setup} records what is judged: the judging paradigm, the task whose outputs the judge assesses, and the languages it is deployed on.
Table~\ref{tab:ann-reliability} records the judge itself --- which models act as judges, what kind of models they are, and how and, crucially, in which languages their reliability is established.
For Table~\ref{tab:ann-reliability} we distinguish \emph{what} a judge is validated against --- human judgments (\textit{human}), gold-standard benchmark labels (\textit{gold}), the judge's own repeated outputs (\textit{self}), or nothing (\xmark) --- from \emph{which} of the
deployment languages that validation covers: \textit{all} of them, a \textit{subset}, English only, or not specified (\textit{n.s.}) when the paper reports an aggregate figure without saying which languages contribute to it.

\section{Checklist for Using LLM Judges}
\label{app:checklist}

The following is a checklist to help design, describe, and review experiments with LLM judges.
We recommend that papers that include such experiments answer all relevant questions.

\paragraph{Screening questions}

\begin{itemize}
    \item Did you use one or more LLMs to evaluate the outputs of other LMs directly, i.e. by feeding model-generated text into a judge model (``LLM-as-evaluator'')?
    \item Did you use one or more LLMs to annotate non-model-generated text (``LLM-as-annotator'')? Reasons could include measuring the LLM's performance on the annotation task, creating a dataset, orevaluating other models indirectly by comparing them the LLM's reference annotation.
    \item Is your primary contribution, or one of your major contributions, to \textit{evaluate} LLM-as-a-Judge (LLM-as-evaluator or LLM-as-annotator)? The submission should clearly state whether the LLM is used as an evaluator, annotator, or both.
    \item Is your primary contribution, or one of your major contributions, to \textit{improve} LLM-as-a-Judge (LLM-as-evaluator, or LLM-as-annotator)?
\end{itemize}

\noindent
If you answered {\bf Yes} to one or more of the questions above, 
please also answer the following questions.

\paragraph{LLM-as-a-Judge}

\begin{itemize}
    \item Did you check if non-LLM-based metrics are equally or more feasible, compared to LLM-as-a-Judge? Did you use them alongside or instead of an LLM judge?
    \item Did you consider the real-world relevance and risks of LLM-as-a-Judge in your particular application domains? If relevant, did you discuss them in the submission?
    \item Did you report performance for multiple different Judge/Evaluator LLMs or an ensemble Judge/Evaluator?
\end{itemize}

\paragraph{Languages}
\begin{itemize}
    \item To which languages did you apply LLM-as-a-Judge? How well-resourced are these languages (e.g., in terms of the scale by \citealp{joshi-etal-2020-state})?
    \item Did you validate the \textit{Judge/Evaluator performance} of your Judge/Evaluator LLMs in each language you applied it to? Does the submission clearly state which languages the Judges were (not) validated on?
    \item Did you validate \textit{general LM or other task-specific performance} (not Judge/Evaluator-specific) of your Judge/Evaluator LLMs in each language you applied it to? Does the submission clearly state which languages the Judges were (not) validated on?
    \item Did you provide statistical confidence estimates for each language?
\end{itemize}

\paragraph{Human annotation}
\begin{itemize}
    \item Did you conduct human evaluation (at least on a subset of the data) for each language?
    \item Did you account for the possibility that humans might disagree on how to judge a given text or may introduce their own biases?
    \item Did you document your selection criteria for human annotators/evaluators, their language competence and domain expertise, as well as any other relevant criteria (academic status, remuneration, conflicts of interest, ...)?
    \item Did you consider representativeness beyond language, such as potential cultural biases or additional representational dimensions and risks associated with image processing?
\end{itemize}

\onecolumn
\begin{landscape}
\footnotesize
\setlength{\tabcolsep}{4pt}
\begin{longtable}{@{}p{3.4cm}cp{8.6cm}p{9.8cm}@{}}
\caption{What is judged, for the 33 in-scope papers. \textbf{Par.} = judging paradigm: DS = direct scoring, MD = multi-dimensional judging,
PC = pairwise comparison and ranking, MS = multi-step/reasoning-based judging, ME = multi-agent/ensemble judging. Papers are ordered by year, then alphabetically.}
\label{tab:ann-setup} \\
\toprule
\textbf{Paper} & \textbf{Par.} & \textbf{Judged task} & \textbf{Languages} \\
\midrule
\endfirsthead
\toprule
\textbf{Paper} & \textbf{Par.} & \textbf{Judged task} & \textbf{Languages} \\
\midrule
\endhead
\bottomrule
\endlastfoot
\citet{boughorbel-hawasly-2023-analyzing} & DS & Multi-turn instruction following and open-ended generation (MT-Bench: writing, extraction, reasoning, math, coding, role-play, humanities, STEM) & Arabic, English \\
\addlinespace[2pt]
\citet{boughorbel-etal-2024-improving} & MD, DS & Arabic short-story completion: the judge scores stories generated by small LMs trained on machine-translated data & Arabic \\
\addlinespace[2pt]
\citet{devine-2024-sure} & PC, DS & Two judging stages: (i) ranking of seven candidate LLM responses to open-ended chat prompts; (ii) scoring of the trained models' MT-Bench chat answers across eight categories & Stage (i): 2,714 prompts across 62 languages. Stage (ii): 6 languages (Chinese, English, French, German, Japanese, Russian) \\
\addlinespace[2pt]
\citet{dinh-etal-2024-sciex} & DS & Grading of free-form answers to scientific exam questions & English, German \\
\addlinespace[2pt]
\citet{forde-etal-2024-evaluating} & PC & Summarization & English, Chinese, Indonesian \\
\addlinespace[2pt]
\citet{gupta-etal-2024-walledeval} & DS & Safety evaluation of model responses & English, Arabic, Filipino, French, Hindi, Russian, Serbian, Spanish \\
\addlinespace[2pt]
\citet{hada-etal-2024-large} & DS & Open prompt, continue-writing and summarization & English, French, German, Spanish, Chinese, Japanese, Italian, Brazilian Portuguese, Czech \\
\addlinespace[2pt]
\citet{hada-etal-2024-metal} & MD & Summarization (XL-Sum) & English, French, Chinese, Hindi, Arabic, Bengali, Russian, Turkish, Japanese, Swahili \\
\addlinespace[2pt]
\citet{liu-etal-2024-optimizing} & MD & Conversational Tutoring Systems (CTS), specifically for L2 (Second Language) learning through an image description task & English--Chinese, English--Korean \\
\addlinespace[2pt]
\citet{marchisio-etal-2024-quantization} & PC & MMLU, Math reasoning (MGSM), machine translation (Flores), language confusion, Aya evaluation (QA) & 22 languages (Arabic, Czech, German, English, Spanish, Greek, Persian, French, Hindi, Indonesian, Italian, Japanese, Korean, Dutch, Polish, Portuguese, Romanian, Russian, Turkish, Ukrainian, Vietnamese, Chinese) \\
\addlinespace[2pt]
\citet{mendonca-etal-2024-ecoh} & DS & Dialog generation, dialog coherence evaluation & English, French, German, Italian, Chinese \\
\addlinespace[2pt]
\citet{qian-etal-2024-large} & DS & Machine translation evaluation & English-German (EN-DE), English-Marathi (EN-MR), English-Chinese (EN-ZH), Estonian-English (ET-EN), Nepali-English (NE-EN), Romanian-English (RO-EN), Russian-English (RU-EN), and Sinhala-English (SI-EN) \\
\addlinespace[2pt]
\citet{raju-etal-2024-constructing} & PC, MS & (1) Determining which LLM responses are preferable over others in a chat setting;
(2) topic/language clustering (used to break down response quality by topic/language) & Japanese, Arabic, Thai, Hungarian, Russian, Serbian, Slovenian, Turkish \\
\addlinespace[2pt]
\citet{sato-etal-2024-tmu} & DS & Machine translation quality estimation  & English $\rightarrow$ Gujarati, Hindi, Tamil, Telugu \\
\addlinespace[2pt]
\citet{watts-etal-2024-pariksha} & PC, DS & Open-ended question answering & Hindi, Tamil, Telugu, Malayalam, Kannada, Marathi, Odia, Bengali, Gujarati, Punjabi \\
\addlinespace[2pt]
\citet{bennie-etal-2025-codeofconduct} & DS & Multilingual counterspeech generation & English, Spanish, Basque, Italian \\
\addlinespace[2pt]
\citet{cruz-blandon-etal-2025-memerag} & DS, MD, MS & Retrieval-augmented generation; meta-evaluation of LLM judges for RAG & English, German, Spanish, French, Hindi \\
\addlinespace[2pt]
\citet{doddapaneni-etal-2025-cross} & MD & Evaluation of instruction-tuned LLM outputs (not broken down beyond broad categories such as reasoning and safety) & Bengali, German, French, Hindi, Telugu, Urdu \\
\addlinespace[2pt]
\citet{duwal-etal-2025-domain} & MD & Domain-Adaptive Continual Pre-training (DAPT), specifically for Nepali generation, instruction following, and linguistic knowledge acquisition (dependency resolution) & Nepali \\
\addlinespace[2pt]
\citet{ellinger-groh-2025-depends} & MD & Resolving referential ambiguity (disambiguation/interpretation) & English, Arabic, French, Chinese, Russian \\
\addlinespace[2pt]
\citet{farhan-2025-hyderabadi} & DS & Multilingual counterspeech generation & English, Basque, Italian, Spanish \\
\addlinespace[2pt]
\citet{fu-liu-2025-reliable} & DS & Question Answering (XQUAD), Math Question Answering (MGSM), Machine Translation (WMT23), Summarization (WikiLingua), Dialogue Generation (XDaily Dialog) & 25 languages (English, German, Russian, Spanish, Chinese, Vietnamese, Turkish, Greek, Romanian, Thai, Hindi, French, Japanese, Swahili, Bengali, Telugu, Czech, Ukrainian, Hebrew, Portuguese, Italian, Indonesian, Dutch, Arabic, Korean) \\
\addlinespace[2pt]
\citet{kocmi-etal-2025-findings-wmt25} & MD, DS & Open-ended generation (OEG), cross-lingual summarization (XLSum), machine translation (MT) & OEG 20 langs: Arabic, Bengali, Czech, German, Greek, English, Farsi, Hindi, Icelandic, Indonesian, Italian, Japanese, Kannada, Korean, Romanian, Russian, Serbian (both Latin and Cyrillic), Ukrainian, Chinese --> 16 of these are included in the human eval; XLSum 14 langs: target languages are Arabic (Egyptian), Czech, Chinese (Simplified), French, German, Hindi, Indonesian, Italian, Japanese, Korean, Russian, Spanish, Swedish, Turkish; MT: source languages: Czech, English, Japanese (32 language pairs total, mostly English->X) en→(ar, bho, bn, cs, de, el, et, fa, hi, id, is, it, ja, kn, ko, lt, mas, mr, ro, ru, sr-Cyrl, sr-Latn, sv, th, tr, uk, vi, zh), ja→zh, cs→de, cs→uk \\
\addlinespace[2pt]
\citet{lu-etal-2025-mqm} & MS & Machine translation quality assessment  & En--De, En--Ru, Zh--En; English $\rightarrow$ Assamese, Maithili, Kannada, Punjabi \\
\addlinespace[2pt]
\citet{marquez-etal-2025-nlp} & MD, PC & Multilingual counterspeech generation & English, Spanish, Basque, Italian \\
\addlinespace[2pt]
\citet{mukherjee-etal-2025-evaluating} & MD & Text style transfer: sentiment transfer (positive to negative statements and vice versa) and detoxification (toxic to clean text) & English, Hindi, Bengali \\
\addlinespace[2pt]
\citet{niculae-etal-2025-dr} & DS & Evaluating and improving the communication quality of medical advice (telemedicine) & Romanian \\
\addlinespace[2pt]
\citet{niklaus-etal-2025-swiltra} & DS, MD, MS, ME & Machine translation  & German, French, Italian, Romansh, English \\
\addlinespace[2pt]
\citet{okewunmi-etal-2025-evaluating} & MD & Question answering & Yor\`ub\'a, English \\
\addlinespace[2pt]
\citet{rolshoven-etal-2025-unlocking} & MD & Cross-lingual legal document summarization (judicial headnote generation) & German, French, Italian \\
\addlinespace[2pt]
\citet{sitaram-etal-2025-multilingual} & MD & Summarization & Arabic, Hebrew, Vietnamese, Indonesian, Latvian, Lithuanian, Welsh, Greek, Danish, Norwegian (Bokmål), Swedish, Dutch (Netherlands/Flemish), English, Standard High German (Germany/Switzerland), Russian, Ukrainian, Czech, Polish, Slovak, Bulgarian, Croatian, Serbian, Slovene, Catalan, French (Metropolitan/Quebec), Italian, Portuguese (Brazil/Portugal), Romanian, Spanish, Thai, Standard Chinese (Simplified/Traditional), Japanese, Korean, Turkish, Estonian, Finnish, and Hungarian \\
\addlinespace[2pt]
\citet{thakur-etal-2025-mirage} & PC & Arena-based Retrieval-Augmented Generation (RAG) evaluation:  where systems compete each other in a tournament and an LLM based teacher is used as the judge for evaluation & 18 languages (Arabic, Bengali, German, English, Spanish, Persian, Finnish, French, Hindi, Indonesian, Japanese, Korean, Russian, Swahili, Telugu, Thai, Yor\`ub\'a, Chinese) \\
\addlinespace[2pt]
\citet{umutlu-etal-2025-evaluating} & MD & Reading comprehension (XQuAD, Belebele), common sense causal reasoning (XCOPA), abstractive summarization (XL-Sum, TR-PLU Summarization), natural language inference (XNLI), goal/step inference (Turkish PLU subsets), Named Entity Recognition (WikiANN), Part of Speech tagging (UDPOS), question answering (MKQA, MMLU-Pro-TR), offensive language detection (OffensEval-TR), and semantic textual similarity (STSb-TR). The LLMs are used to judge the quality of the dataset not directly working on these tasks. & Turkish \\
\addlinespace[2pt]
\end{longtable}
\end{landscape}

\begin{landscape}
\footnotesize
\setlength{\tabcolsep}{4pt}
\begin{longtable}{@{}p{2.8cm}p{4.0cm}cccp{5.1cm}p{7.6cm}@{}}
\caption{The judge and its reliability, per paper.
\textbf{Val.} = what the judge is validated against (\textit{human} = human judgments or expert annotation, \textit{gold} = gold-standard
benchmark labels, \textit{self} = internal consistency only, \xmark\ = not validated). \textbf{Lang.} = how much of the paper's own
language coverage that validation extends to (\textit{all}, a \textit{subset}, English only, or \textit{n.s.} = the paper reports an
aggregate figure without specifying the languages). \textbf{Pool} characterises the roster: \textit{closed} = closed-source generalist models
only, \textit{open} = open-weight generalist models only, \textit{spec.} = judge-tuned or domain-fine-tuned evaluator models only, \textit{mixed} = models from more than one of these families.
Papers are ordered by year, then alphabetically.}
\label{tab:ann-reliability} \\
\toprule
\textbf{Paper} & \textbf{Judge model(s)} & \textbf{Pool} & \textbf{Val.} & \textbf{Lang.} & \textbf{How reliability is established} & \textbf{Reported reliability, and what it shows about languages} \\
\midrule
\endfirsthead
\toprule
\textbf{Paper} & \textbf{Judge model(s)} & \textbf{Pool} & \textbf{Val.} & \textbf{Lang.} & \textbf{How reliability is established} & \textbf{Reported reliability, and what it shows about languages} \\
\midrule
\endhead
\bottomrule
\endlastfoot
\citet{boughorbel-hawasly-2023-analyzing} & GPT-4 & closed & \xmark & -- & No comparison against human judgments or gold labels. The judge is checked only in a qualitative \emph{Score consistency} analysis: GPT-4's judgments are clustered by numerical rating and GPT-4 is then asked to summarise its own justification texts for each score level. No numeric consistency statistic is reported. & The consistency analysis is explicitly qualitative and yields no agreement figure; the authors conclude that ``the justifications are reasonably consistent across models and categories, indicating an acceptable level of impartiality.'' The quantitative reliability claim is inherited from English MT-Bench, where GPT-4 is reported to agree with human judges ca.\ 85\% of the time. The consistency check pools judgments across models and categories and reports no per-language statistic. No Arabic judgment is compared against a human or gold reference. \\
\addlinespace[2pt]
\citet{boughorbel-etal-2024-improving} & GPT-4 & closed & \xmark & -- & The judge is not directly evaluated. Model outputs are additionally inspected with sparse-autoencoder dictionary learning and token set enrichment analysis, and top results are manually checked. & The authors report qualitatively that evaluating consistency, grammar and creativity is sometimes challenging for GPT-4 and note its limitations in complex Arabic evaluation compared to English, but do not quantify this. Arabic reliability is acknowledged as a concern in prose but never measured; the underlying methodology was validated only on English. \\
\addlinespace[2pt]
\citet{devine-2024-sure} & GPT-4 & closed & self & n.s. & No comparison against human judgments or gold labels. Reliability is assessed only as self-consistency: Kendall's $W$ over five repeated rankings of the same responses with randomised orderings. & The judge is markedly inconsistent with itself: only 8.4\% of top-ranked and 20.2\% of bottom-ranked responses receive the same rank in all five trials. Kendall's $W$ is reported only as an aggregate over the whole dataset, with no per-language breakdown. \\
\addlinespace[2pt]
\citet{dinh-etal-2024-sciex} & Mixtral, Llama-3, GPT-4V & mixed & human & all & Comparison against grades assigned by human expert graders (Pearson correlation). & Strong agreement overall: 0.948 Pearson correlation with expert grades, which the authors consider reliable enough to replace human experts in future evaluations. The correlation with expert grades is reported pooled over both languages. \\
\addlinespace[2pt]
\citet{forde-etal-2024-evaluating} & GPT-4 & closed & human & all & reliability of automatic metrics is established relative to an aggregate human preference benchmark & Stronger link between automated metrics and human judgments for English (than Chinese and Indonesian). However, there is not an independent reliability of human judgments. \\
\addlinespace[2pt]
\citet{gupta-etal-2024-walledeval} & LlamaGuard v1/2/3, WalledGuard and WalledGuard-Adv (the authors' own safety classifiers); the toolkit also supports generic judges such as GPT-4 and Claude & mixed & gold & all & Judges are evaluated by comparing their predictions against gold-standard labels. Metrics include the percentage of correct classifications (accuracy) on safe and unsafe prompts/responses. & Performance varies by language; for instance, judges find it harder to classify samples in the HIXSTEST (Hindi) dataset compared to English. The WalledGuard Adv can reach 92.81\% acc on English. \\
\addlinespace[2pt]
\citet{hada-etal-2024-large} & GPT-4 & closed & human & all & Calibration against human judgements: IAA (3 annotators ) and GPT. & LM-based evaluators may perform worse on low-resource and non-Latin script languages. But it's mostly consistent for high-resource languages. \\
\addlinespace[2pt]
\citet{hada-etal-2024-metal} & GPT-3.5, GPT-4, PaLM2 & closed & human & all & They measured the pairwise agreement between the LLM evaluators and human aggregate scores per language and metric. For ``human scores'', they average the pairwise F1 scores of all the annotators. & Human have better score, GPT-4 with detailed instructions comes closest to human evaluation. Performance is worse on low-resource languages like Bengali. \\
\addlinespace[2pt]
\citet{liu-etal-2024-optimizing} & GPT-4,Gpt-3,5, Llama, Mistral  & mixed & human & all & The judge's performance is benchmarked against human experts. Reliability is measured by comparing the judge’s predictions to the labels provided by human experts using Accuracy and F1-score. & Strong agreement with expert human annotations (higher agreement for Chinese–English than Korean–English). \\
\addlinespace[2pt]
\citet{marchisio-etal-2024-quantization} & GPT-4 & closed & human & subset & They used human evaluation to confirm LLM-as-a-judge, they claim it paints a similar picture. Humans used a 5-point Likert scale, then express a preference (tie, weak preference, strong preference). & It perform similarly to human evaluation, although many of the languages are rather mid-resource to high-resource, used in evaluation. \\
\addlinespace[2pt]
\citet{mendonca-etal-2024-ecoh} & Proposed model "ECoh", based on Qwen1.5-Chat finetuned in various sizes from 0.5B to 72B, on English-only or on data from all languages (ml); Baselines: RoBERTa-large, UniEval, Qwen1.5-Chat zero-shot and one-shot, GPT-3.5-Turbo & mixed & human & English-only & Main results are scored against \emph{LLM-generated} labels (GPT-3.5-turbo for the development set, GPT-4 for the test set), using point-biserial correlation, F1, and BLEU against the GPT-4 explanation; GPT-4 additionally acts as a judge-of-judge for explanation quality. The only comparison against \emph{human} judgements is on FED-turn, an existing English-only benchmark (Pearson $r$ with Relevance and Overall ratings). A separate 100-sample check by one expert linguist per language validates the GPT-4-\emph{generated data}, not the judge's predictions. & On one dataset (DailyDialog), ECoh-multilingual-4B outperforms all other by a relatively clear margin in the metrics point biserial correlation, F1-score, BLEU. Second come ECoh-en-4B and then smaller sizes of ECoh-en and -ml as well as GPT-3.5 and 1-shot Qwen. The GPT-4 judge-of-judge metric slightly prefers GPT-3.5 over ECoh and ECoh-en over ECoh-ml. On the other dataset (Persona), the correlation and F1 trends are very similar, but BLEU prefers the smallest (0.5B) ECoh variant, and performance on the multilingual subset is generally better than on the French-only subset. Pearson correlation with human scores from FED-turn dataset ranks differently: GPT-4 > GPT-3.5 > 1-shot Qwen-7B. ECoh mostly ranks at the bottom, together with the RoBERTa and UniEval baselines. \\
\addlinespace[2pt]
\citet{qian-etal-2024-large} & Llama-2-7B, Llama-2-13B, Gemma-7B, OpenChat-3.5, Qwen1.5-14B, Mixtral-8x7B-AWQ & open & human & all & The predicted scores are evaluated for reliability and accuracy by calculating the Spearman correlation between the LLM's predicted scores and the human-annotated ``gold-standard'' DA scores. & LLM performance generally lags behind fine-tuned encoder-based models (like TransQuest). \\
\addlinespace[2pt]
\citet{raju-etal-2024-constructing} & GPT-4o, GPT-4o-mini, Llama-3.1-405B-Instruct, Llama-3.1-70B-Instruct-Turbo & mixed & human & n.s. & Chatbots (base task that is evaluated with LLM judges) are ranked by winrate and ELO, this is further broken down by user prompt topic and language. The rankings produced by the four LLM evaluator judges are compared on Spearman correlation coefficient, Model separability, Agreement with Confidence Interval, Brier Score.
 & Worse chatbot separability (84.44\% vs 100\%) than ChatbotArena, which uses human judges only; roughly on-par (slightly better) chatbot separability with/than ArenaHard v0.1 (80\%) and AlpacaEval v2.0 LC (73.33\%), which use LLM judges; much better alignment with ChatbotArena than ArenaHard or AlpacaEval, both in terms of ``Agreement with CI'' and Spearman; evaluation prompt was engineered / adjusted from Zheng et al., trying to avoid the following two issues that were observed in development: (1) judges tended to prefer English responses, even if user asked for specific non-English language; (2) for coding questions, judges tended to prefer lengthy explanations over code quality. \\
\addlinespace[2pt]
\citet{sato-etal-2024-tmu} & GPT-4o-mini (off-the-shelf and fine-tuned) & closed & human & all & Spearman's rank correlation with human annotations. & Low to moderate correlation, with $\rho$ between 0.27 and 0.52. \\
\addlinespace[2pt]
\citet{watts-etal-2024-pariksha} & GPT-4-32K & closed & human & all & Percentage agreement and Fleiss' $\kappa$ between human and LLM judgments, plus Kendall's $\tau$ between the human-derived and LLM-derived leaderboards. & Humans and LLMs agree fairly well in the pairwise setting but the agreement drops for direct assessment evaluation especially for languages such as Bengali and Odia. Average Fleiss $\kappa$ correlation between Human-LLM is 0.49 for pairwise comparison. The LLM evaluator has very low agreement with humans on Marathi, Bengali and Punjabi. Average Fleiss $\kappa$ correlations between Human-LLM is 0.31 in direct assessment. The human-LLM agreement is significantly lower particularly for Bengali and Odia. Regarding the agreement between the leaderboards, a high Kendall $\tau$ score of  0.76 was obtained for the Elo leaderboards, which signifies that the human and LLM-evaluator agree on the general trends. \\
\addlinespace[2pt]
\citet{bennie-etal-2025-codeofconduct} & JudgeLM (a fine-tuned Vicuna-based model) + JudgeEUS & spec. & self & all & No comparison against human judgment. Internal evaluation (pairwise, order-swapped judging, score consistency checks, and comparison with the shared-task scoring procedure)  & JudgeLM is useful for ranking and optimizing counterspeech generation but its reliability is not clear across languages. In addition, Judge-EUS evaluator was used for Basque which may lead to cross-language inconsistencies for evaluation. \\
\addlinespace[2pt]
\citet{cruz-blandon-etal-2025-memerag} & GPT-4o-mini, Qwen2.5-32B, Llama3.2-90B, Llama3.2-11B & mixed & human & all & Balanced accuracy against human labels. & Highly language-specific: on English, GPT-4o performs best with Llama and Qwen 6-8 pp. behind. On the other European languages, Qwen performs best with GPT-4o closely behind (-1-3 pp) and Llama much worse (-8-16 pp). On Hindi, Qwen and GPT-4o still perform well, but Llama3.2 9B does, too. \\
\addlinespace[2pt]
\citet{doddapaneni-etal-2025-cross} & Llama-3.1-8B-I (``Hercule'') fine-tuned for judging outputs in the relevant languages, once per language, and also once jointly for all 6 languages, and also some model-merging experiments; Llama-3.1-8B (``Prometheus'') fine-tuned for judging outputs in English (different dataset); Gemma, Sarvam, Aya23 fine-tuned for judging outputs in the relevant languages; zero-shot Llama-3.1-405B-I, GPT-4o, Gemini-1.5-Pro & mixed & human & all & Kappa with GPT-4o scores, pearson's r correlation with human scores. & Depends on the model and to some extent on the language; fine-tuned models typically beat zero-shot judges even if the latter are much larger; best judge (theirs) that has access to reference answers: kappa between 0.69-0.75, pearson's r between 0.42-0.78; without reference answers, the performance is worse (kappa between 0.63-0.68); when the reference is in the language of the task, the performance is also worse (kappa between 0.59-0.73 / no results); the judge trained jointly on all 6 languages is worse than the individual language-specific evaluators. \\
\addlinespace[2pt]
\citet{duwal-etal-2025-domain} & GPT-4o & closed & \xmark & -- & The judge is not evaluated against a human or gold-standard reference. & While not formally benchmarked, the authors justified its use by citing the ``multilinguality of frontier language models'' and their proven performance in languages they do not officially support, including Nepali. \\
\addlinespace[2pt]
\citet{ellinger-groh-2025-depends} & GPT 4.1 & closed & human & English-only & Human annotations as gold standard (Kendall, Spearman, accuracy).   & Moderately strong alignment with human judgements (98\% agreement for classification and 97.8\% exact-match accuracy for entity extraction).  \\
\addlinespace[2pt]
\citet{farhan-2025-hyderabadi} & JudgeLM (a fine-tuned Vicuna-based model) & spec. & \xmark & n.s. & No human evaluation of the judge is carried out. & JudgeLM is applied to different languages but the reliability is not reported. \\
\addlinespace[2pt]
\citet{fu-liu-2025-reliable} & Gemini-2.0-Flash GPT-3.5-turbo, GPT-4o-2024-08-06, Llama-3.3-70b and Qwen-2.5-72b, Aya-Expanse-32b & mixed & self & all & Measured using Fleiss' Kappa (FK) to determine inter-rater agreement across different languages for the same parallel content. & LLMs struggle to achieve consistent results across languages, with an average Fleiss' Kappa of approximately 0.3. \\
\addlinespace[2pt]
\citet{kocmi-etal-2025-findings-wmt25} & MT (not evaluated; incidental in composite score): GPT-4.1, Command A; MT (evaluated; best to worst): GPT-4.1, Claude 4, Command A, DeepSeek V3, Qwen3-235B, AyaExpanse-32B, Llama 4 Maverick, Qwen2.5-7B, Llama 3.1-8B, CommandR-7B, AyaExpanse-8B, Mistral-7B; OEG (best to worst): Claude 4, GPT-4.1, Command A, Qwen3-235B, Mistral Medium, DeepSeek V3, Llama 4 Maverick, AyaExpanse-32B, Qwen2.5-7B, Llama 3.1-8B, CommandR-7B, AyaExpanse-8B, Mistral-7B; XLSum (best to worst): GPT-4.1, Command A, Mistral Medium, Llama 4 Maverick, Qwen3-235B, DeepSeek V3, CommandR-7B, Qwen2.5-7B, AyaExpanse-32B, Llama 3.1-8B, AyaExpanse-8B, Mistral-7B. & mixed & human & subset & Correlation against the shared task's human evaluation. & OEG: when ranking different systems: very high correlation for some models ($\sim$ 100B params), for others rather low.
Instance-level correlations are moderate, and the best judge for each criterion varies; XLSum: system-level ranking works very well when done by the larger judge models, for the others the correlation is still relatively high. Instance-level correlations are moderate to low, and the best judge for each criterion varies; MT: ranking scores are not as high as for OEG and XLSum (and again, the bigger models are better). Instance-level scores show low to moderate correlation. \\
\addlinespace[2pt]
\citet{lu-etal-2025-mqm} & Llama-3-8B/70B-Instruct, Mixtral-8x7B/8x22B-Instruct, Qwen1.5-14B/72B-Chat, TowerInstruct-7B/13B & open & human & all &  Pairwise accuracy was used to assess how well evaluator judgments align with human-annotated MQM score. & MQM scores: system-level accuracy ranges from 83\%-88\% across different evaluators; segment-level accuracy ranges from $\sim$ 51\%-55\%; error span quality: SP (Span Precision) ranges from 10-21; MP (Major Errors Precision) ranges from $\sim$ 3-12; MQM-APE filters non-impactful errors and improves reliability and interpretability, but it does not align the error distribution with human evaluation. \\
\addlinespace[2pt]
\citet{marquez-etal-2025-nlp} & JudgeLM (a fine-tuned Vicuna-based model) & spec. & \xmark & -- & The judge is not evaluated. &  The authors observe a divergence between the JudgeLM ranking and the automated metrics (RougeL, BLEU, BERTScore). \\
\addlinespace[2pt]
\citet{mukherjee-etal-2025-evaluating} & GPT-4, Llama-3.1-8B & mixed & human & all & Run correlation with human judgment using Pearson, Spearman, and Kendall’s Tau. Evaluated separately for each dimension. & GPT-4 consistently aligns well with human assessments of overall quality in both Sentiment Transfer and Detoxification. Llama, however, shows weaker correlations. \\
\addlinespace[2pt]
\citet{niculae-etal-2025-dr} & Gemma-12B, Gemma-27B, MedGemma-27B & mixed & human & all & Evaluated using F1-score and Accuracy by comparing LLM judgments against 100 expert-annotated doctor-patient pairs. & The paper highlights that ``traditional'' unoptimized judges do not work well for specialized domains like Romanian telemedicine, whereas optimized versions do. \\
\addlinespace[2pt]
\citet{niklaus-etal-2025-swiltra} & GPT-4o, GPT-4o-mini, Gemini-1.5-pro, Gemini-1.5-flash; both as single models and ensembles & closed & human & n.s. & Spearman correlation, RMSE, and MAE, with human judgments with cross validation; Spearman (Bootstrap). & Highest Spearman correlation with human judgments, compared to METEOR, BERTScore, GEMBA-MQM, BLEURT, and XCOMET. \\
\addlinespace[2pt]
\citet{okewunmi-etal-2025-evaluating} & GPT4.0, Gemini 2.0, Claude 3.7 (Claude 3.7 Sonnet) & closed & \xmark & -- & No evaluation for judging LLMs output (no human evaluation). & Not measured.  \\
\addlinespace[2pt]
\citet{rolshoven-etal-2025-unlocking} & DeepSeek-V3 & open & human & n.s. & Correlation with human expert judgments (lawyers) using the same rubric; Spearman correlation analysis. & Overall JUDGE score shows moderate correlation with human ratings (0.26 Spearman); strongest agreement (0.27-0.30) on concrete, verifiable aspects of legal headnotes (articles, considerations), weaker (0.07) on subjective dimensions (clarity, coherence). \\
\addlinespace[2pt]
\citet{sitaram-etal-2025-multilingual} & GPT-4o, Phi-3.5-MoE-Instruct, Qwen2.5-VL-7B & mixed & human & subset & Agreement with human judgments (only for GPT-4o) with Cohen's $\kappa$. & GPT-4o–human agreement was moderate for gender assumption ($\kappa$ = 0.42) but low for gender mistake ($\kappa$ = 0.14) and quality ($\kappa$ = 0.03) and weaker agreement in some low-resource languages \\
\addlinespace[2pt]
\citet{thakur-etal-2025-mirage} & GPT-4o performs pairwise preference judgments between RAG responses; a surrogate judge (a non-LLM random forest model; computationally less expensive) is trained, which achieves Kendall-Tau = 0.909 average correlation with GPT-4o; Llama-3 (8B) model is used as a judge to evaluate the fluency of a RAG response and the answer overlap.  & mixed & \xmark & -- &  Not evaluated against human or gold-label references. & Not assessed. \\
\addlinespace[2pt]
\citet{umutlu-etal-2025-evaluating} & Llama-3.3-70B-Instruct, Gemma-2-27B-it, and GPT-4o & mixed & human & all & The LLMs are analyzed for their labeling performance compared to human annotations. & LLM judges are generally less effective than human annotators, especially in understanding cultural common sense and interpreting fluency/ambiguity. They have less than 80\% overlap with human annotations for all criteria except technical term usage. GPT-4o performed better in grammatical and technical tasks, while Llama-3.3-70B was superior in correctness and cultural knowledge evaluation. \\
\addlinespace[2pt]
\end{longtable}
\end{landscape}
\twocolumn

\end{document}